\documentclass[runningheads]{llncs}
\usepackage[T1]{fontenc}
\usepackage{graphicx}
\usepackage{amsmath}
\usepackage{algorithm}
\usepackage{algpseudocode}
\usepackage{geometry}
\begin{document}
\title{Solving the Pod Repositioning Problem with Deep Reinforced Adaptive Large Neighborhood Search}
%
%
\author{Lin Xie\orcidID{0000-0002-3168-4922}, Hanyi Li}
\authorrunning{L. Xie et al.}
%
\institute{Brandenburg University of Technology Cottbus-Senftenberg, Cottbus, D-03046, Germany \\
\email{lin.xie@b-tu.de}}
\maketitle              
\begin{abstract}
The Pod Repositioning Problem (PRP) in Robotic Mobile Fulfillment Systems (RMFS) involves selecting optimal storage locations for pods returning from pick stations. This work presents an improved solution method that integrates Adaptive Large Neighborhood Search (ALNS) with Deep Reinforcement Learning (DRL). A DRL agent dynamically selects destroy and repair operators and adjusts key parameters such as destruction degree and acceptance thresholds during the search. Specialized heuristics for both operators are designed to reflect PRP-specific characteristics, including pod usage frequency and movement costs. Computational results show that this DRL-guided ALNS outperforms traditional approaches such as cheapest-place, fixed-place, binary integer programming, and static heuristics. The method demonstrates strong solution quality and illustrating the benefit of learning-driven control within combinatorial optimization for warehouse systems.

\keywords{Pod repositioning problem  \and Deep reinforcement learning \and Adaptive large neightborhood search \and Warehouse logistics}
\end{abstract}
\section{Introduction}

Robotic Mobile Fulfillment Systems (RMFS) have revolutionized warehouse operations by enabling mobile robots to transport storage pods (also known as racks) between storage areas and operational stations such as picking or replenishment stations. A critical operational decision in RMFS is determining the optimal storage location for a pod after it has been processed at a station. This decision, referred to as the \textit{Pod Repositioning Problem (PRP)} or \textit{Rack Storage Assignment Problem}, significantly impacts robot travel distances, inventory accessibility, and overall system throughput. For example, pods that are not expected to be needed soon should be placed farther from stations to preserve nearby storage locations for high-frequency or urgent items.

The PRP has been formally defined in a deterministic context by Krenzler et al.~\cite{Krenzler2018}, who proposed a sequential decision-making model aimed at minimizing cumulative travel costs associated with pod storage assignments. Their work evaluated various strategies, including binary integer programming, cheapest-place heuristics, fixed-place policies, and a tetris-like heuristic. Similarly, Weidinger et al.~\cite{Weidinger2018} addressed the similar storage assignment problem, formulating it as an interval scheduling problem and employing mixed-integer programming to optimize pod movements and workload balancing.

Further contributions to deterministic storage assignment strategies include Merschformann et al.~\cite{Merschformann2018} considered both active and passive repositioning scenarios. Yuan et al.~\cite{Yuan2019} developed a velocity-based storage assignment policy for semi-automated storage systems, demonstrating the benefits of class-based storage over random assignment. Zhuang et al.~\cite{Zhuang2022} investigated the pod storage and automated guided vehicle (AGV) task assignment problem, proposing a meta-heuristic decomposition approach to optimize pod retrieval and repositioning.

While these deterministic approaches have proven effective in structured environments, they often rely on static heuristics or require extensive parameter tuning. To enhance adaptability and performance, metaheuristic frameworks like Adaptive Large Neighborhood Search (ALNS)~\cite{Ropke2006} offer flexible solution methods by iteratively destroying and repairing parts of a solution to explore the search space effectively. Reijnen et al.~\cite{Reijnen2024} extended this concept by integrating Deep Reinforcement Learning (DRL) into the ALNS framework, allowing dynamic selection and configuration of heuristics during the search process. Although this DRL-guided ALNS (DR-ALNS) has shown promise in routing problems, its application to storage assignment in RMFS remains unexplored.

In parallel, DRL has been applied to dynamic and stochastic variants of warehouse storage problems. Rimélé et al.~\cite{Rimele2021} trained a deep Q-network to learn dynamic storage policies in simulated RMFS environments, while Teck et al.~\cite{Teck2025} proposed a DRL-based method for real-time inventory pod storage and replenishment under uncertainty. These studies emphasize reactive decision-making in stochastic settings. DRL has shown promising results in warehouse operations, including order batching~\cite{Cals2021} and picker routing~\cite{Luttmann2024}, \cite{Luttmann2025}.

Despite the advancements in dynamic and stochastic approaches, many warehouse operations still function under deterministic or semi-deterministic conditions, where order streams and system states are predictable over short horizons. Leveraging deterministic models allows for exploiting known demand patterns and system structures, facilitating more efficient and reliable storage assignment strategies. Building upon the deterministic framework established by Krenzler et al.~\cite{Krenzler2018}, this study aims to enhance storage assignment performance through adaptive metaheuristic methods without introducing stochastic complexities.

\paragraph{Research Gap and Contribution.}
To the best of our knowledge, the integration of DRL into metaheuristic frameworks like ALNS has not been applied to the deterministic PRP in RMFS. Existing DRL approaches focus on dynamic, real-time decision-making, while traditional deterministic methods rely on static heuristics.

This paper makes the following contributions:
\begin{itemize}
    \item We adapt the DR-ALNS framework to the deterministic pod repositioning problem, demonstrating its applicability and effectiveness in structured warehouse environments.
    \item We develop and formalize domain-specific destroy and repair heuristics tailored to the PRP, enhancing the flexibility and performance of the ALNS framework.
    \item Through computational experiments, we compare our approach against established baselines, including cheapest-place, fixed-place, and binary integer programming methods, showcasing improved solution quality and adaptability.
\end{itemize}

This work complements the existing literature by demonstrating that adaptive metaheuristic search, guided by reinforcement learning, can yield substantial improvements even within deterministic, rule-driven storage environments.

\section{Problem Description: Deterministic Pod Repositioning Problem}

We follow the deterministic formulation of the Pod Repositioning Problem (PRP) proposed by Krenzler et al.~\cite{Krenzler2018}, where the goal is to assign returning pods to storage locations over time to minimize total robot travel costs. Each placement decision affects future availability and retrieval efficiency due to the shared and time-constrained nature of storage locations.

Let $T$ denote the planning horizon and $S$ the set of storage locations. At each time step $t \in \{0, \dots, T-1\}$, a set of pods $\mathcal{P}_t$ returns to storage. For each pod $p \in \mathcal{P}_t$, we assign a storage location $a_t(p) \in S$. Let $d(p)$ be the next departure time of $p$, and define the travel costs:
\begin{itemize}
    \item $c_{\text{store}}(p, s, t)$: cost to store pod $p$ at location $s$ at time $t$,
    \item $c_{\text{retrieve}}(p, s, d(p))$: cost to retrieve pod $p$ from location $s$ at time $d(p)$.
\end{itemize}

\subsection*{Objective Function}

The objective is to minimize the total storage and retrieval costs across all time steps:
\[
\min \sum_{t=0}^{T-1} \sum_{p \in \mathcal{P}_t} \left[ c_{\text{store}}(p, a_t(p), t) + c_{\text{retrieve}}(p, a_t(p), d(p)) \right]
\]

\subsection*{Constraints}

\begin{enumerate}
    \item \textbf{No overlap:} A storage location cannot be used by multiple pods during overlapping occupancy periods:
    \[
    \forall s \in S, \quad \text{at most one } p \text{ such that } a_t(p) = s \text{ and } t < d(p)
    \]
    
    \item \textbf{Feasibility:} Each pod must be assigned to an available location at its return time:
    \[
    a_t(p) \in S \quad \text{and } s = a_t(p) \text{ must be unoccupied for } [t, d(p))
    \]
\end{enumerate}

\subsection*{Heuristic Consideration}

In practice, it is beneficial to prioritize frequently used pods for locations that are closer to pick stations to reduce average retrieval costs. Let $f(p)$ denote the access frequency of pod $p$ and $S_{\text{near}} \subseteq S$ denote the set of preferred (i.e., near-access) storage locations. A heuristic strategy may enforce:
\[
\text{if } f(p) \text{ is high, then prefer } a_t(p) \in S_{\text{near}}
\]
This prioritization is often embedded in the design of repair heuristics (e.g., ABC-based placement).

\subsection*{Discussion}

The PRP is a dynamic, time-indexed assignment problem where the cost and feasibility of each decision are interdependent. Although the system evolves deterministically, its constrained nature creates long-range dependencies between placement decisions. This motivates the use of metaheuristics such as ALNS, which can iteratively refine global placement patterns, and learning-based extensions like DRL, which can adaptively control heuristic behavior in context.

\section{Adaptive Large Neighborhood Search for the Pod Repositioning Problem}

This section details our implementation of ALNS, including the destroy and repair heuristics, solution feasibility and acceptance criteria.

\subsection{ALNS Framework}

ALNS iteratively refines an initial solution through a cycle of destruction and repair operations. A set of heuristics for each phase is maintained, with adaptive weight updates promoting those that consistently lead to improved solutions. The general process is shown in Algorithm~\ref{alg:alns}.

\begin{algorithm}[H]
\caption{Adaptive Large Neighborhood Search}
\label{alg:alns}
\begin{algorithmic}[1]
\State \textbf{Input:} Initial solution $s_0$, temperature schedule $(T_{start}, T_{stop}, \alpha)$, Markov chain length $M$, acceptance probability $p_{accept}$, heuristic weights $w_d$, $w_r$
\State Set $s \gets s_0$, $s^* \gets s_0$
\For{$i = 1$ to $M$}
    \State Select destroy heuristic $d \in \mathcal{D}$ with probability proportional to $w_d$
    \State $s' \gets d(s)$
    \State Select repair heuristic $r \in \mathcal{R}$ with probability proportional to $w_r$
    \State $s'' \gets r(s')$
    \If{$\text{cost}(s'') < \text{cost}(s^*)$}
        \State $s^* \gets s''$
        \State Accept $s \gets s''$ unconditionally
        \State Update $w_d, w_r$ (global improvement)
    \ElsIf{$\text{cost}(s'') < \text{cost}(s)$ or rand() $< p_{accept}$}
        \State Accept $s \gets s''$
        \State Update $w_d, w_r$ (local or accepted worse solution)
    \Else
        \State Update $w_d, w_r$ (no improvement)
    \EndIf
    \State Update temperature $T \gets \max(T \cdot \alpha, T_{stop})$
\EndFor
\State \Return $s^*$
\end{algorithmic}
\end{algorithm}

To guide the selection of destroy and repair heuristics, the ALNS framework maintains a set of weights $w_d$ and $w_r$ associated with each heuristic. At the beginning of each iteration, heuristics are chosen probabilistically according to these weights. After the resulting solution is evaluated, the weights are updated using a score-based reward scheme inspired by the original ALNS method~\cite{Ropke2006}.

Let $\sigma \in \{0, 1, 2, 3\}$ be the reward assigned based on the outcome of an iteration:
\begin{itemize}
    \item $\sigma = 3$ if the new solution improves the global best (global improvement),
    \item $\sigma = 2$ if it improves the current solution (local improvement),
    \item $\sigma = 1$ if it is accepted but not better (accepted worse),
    \item $\sigma = 0$ if it is rejected.
\end{itemize}

The weight of a selected heuristic $h$ is then updated as:
\[
w_h \leftarrow \lambda \cdot w_h + (1 - \lambda) \cdot \sigma
\]
where $\lambda \in [0,1]$ is a reaction factor controlling how strongly new feedback influences the weight. This adaptive mechanism enables the algorithm to favor heuristics that consistently lead to better solutions, while still allowing underperforming heuristics occasional influence for exploration.

The performance of ALNS critically depends on the design of effective \textit{destroy} and \textit{repair} heuristics. In this section, we expand on the heuristics implemented in our approach to the PRP, including their rationale, working principles, and pseudocode.

\subsection{Destroy Heuristics}

Destroy heuristics remove a subset of the pod-to-location assignments from the current solution $s$ over a sequence of iterations. This creates space for exploring alternative placements and escaping local optima. Let $I_{\text{destroy}} \subseteq \{0, \dots, N-1\}$ be the set of iterations selected for destruction.

\subsubsection*{Random Destruction}

This heuristic selects a consecutive range of $k = \lfloor \texttt{DoD} \cdot N \rfloor$ iterations uniformly at random and removes the pod assignments in those iterations. It encourages exploration by applying changes across arbitrary regions of the solution. We use a parameterized Degree of Destruction (DoD)\footnote{The Degree of Destruction (DoD) is a scalar in $(0,1]$ that determines how many iterations are selected for removal in the destroy phase. If $N$ is the total number of iterations, then $\lfloor \texttt{DoD} \cdot N \rfloor$ entries are destroyed. Higher DoD values increase exploration, while lower values favor local refinement.} to control the intensity of each destroy operation.

\begin{algorithm}[H]
\caption{Random Destruction}
\begin{algorithmic}[1]
\Function{DestroyRandom}{$s, \texttt{DoD}$}
    \State $k \gets \lfloor \texttt{DoD} \cdot N \rfloor$
    \State Choose $i_{\text{start}} \in \{0, \dots, N-k\}$ uniformly
    \State $I_{\text{destroy}} \gets \{i_{\text{start}}, \dots, i_{\text{start}}+k-1\}$
    \For{$i \in I_{\text{destroy}}$}
        \State Remove pod assignment in iteration $i$ from $s$
    \EndFor
    \State \Return $s'$
\EndFunction
\end{algorithmic}
\end{algorithm}

\subsubsection*{High-Cost Destruction}

This heuristic targets the most expensive region of the solution. It identifies a window of $k$ consecutive iterations that cumulatively contribute the highest cost, and removes their assignments. It supports intensification by attacking problematic areas.

\begin{algorithm}[H]
\caption{High-Cost Destruction}
\begin{algorithmic}[1]
\Function{DestroyHighCost}{$s, \texttt{DoD}$}
    \State $k \gets \lfloor \texttt{DoD} \cdot N \rfloor$
    \For{$i = 0$ to $N-k$}
        \State $C_i \gets \sum_{j=0}^{k-1} \text{cost}(s[i+j])$
    \EndFor
    \State $i^* \gets \arg\max_i C_i$
    \State $I_{\text{destroy}} \gets \{i^*, \dots, i^*+k-1\}$
    \For{$i \in I_{\text{destroy}}$}
        \State Remove pod assignment in iteration $i$
    \EndFor
    \State \Return $s'$
\EndFunction
\end{algorithmic}
\end{algorithm}

\subsection{Repair Heuristics}

After destruction, repair heuristics reconstruct the partial solution $s'$ by assigning storage placements to the unassigned pods at destroyed iterations $I_{\text{destroy}}$. All heuristics rely on the \texttt{FeasibleLocations} function from Section~3.4 to ensure valid placements.

\subsubsection*{Tetris-Inspired Repair}

This repair heuristic is loosely inspired by the Tetris algorithm from Krenzler et al.~\cite{Krenzler2018}, which places returning pods greedily in time order by selecting the lowest-cost feasible location at each step. In contrast, our approach introduces a two-phase structure to improve placement quality.

In Phase 1, we target iterations with the highest placement costs—those likely to create inefficiencies if handled later—and assign them greedily to feasible locations. This anticipates future congestion and mimics the idea of placing the "bulkiest" pieces first, analogous to challenging blocks in Tetris. In Phase 2, the remaining iterations are sorted by pod frequency and urgency (departure time), and are assigned to cost-effective locations using the same greedy placement rule.

\begin{algorithm}[H]
\caption{Tetris-Inspired Repair}
\begin{algorithmic}[1]
\Function{RepairTetris}{$s', I_{\text{destroy}}$}
    \State $U \gets I_{\text{destroy}}$
    \State Sort $U$ by descending placement cost
    \For{$i \in U$}
        \State $P_i \gets \texttt{FeasibleLocations}(i)$
        \If{$P_i \neq \emptyset$}
            \State Assign pod in $i$ to highest-cost location in $P_i$
            \State Remove $i$ from $U$
        \EndIf
    \EndFor
    \State Sort $U$ by pod frequency and next departure time
    \For{$i \in U$}
        \State $P_i \gets \texttt{FeasibleLocations}(i)$
        \If{$P_i \neq \emptyset$}
            \State Assign pod in $i$ to lowest-cost location in $P_i$
        \EndIf
    \EndFor
    \State \Return $s''$
\EndFunction
\end{algorithmic}
\end{algorithm}

\subsubsection*{ABC Priority Repair}

Pods are categorized into three classes based on their usage frequency: Class~A (high frequency, accounting for 70\% of total usage), Class~B (medium frequency, 20\%), and Class~C (low frequency, 10\%). Class~A pods are assigned to the most optimal storage locations, Class~B to moderately favorable ones, and Class~C to the least optimal, reflecting their relative operational importance.

\begin{algorithm}[H]
\caption{ABC Priority Repair}
\begin{algorithmic}[1]
\Function{RepairABC}{$s', I_{\text{destroy}}, \text{frequency}$}
    \For{$i \in I_{\text{destroy}}$}
        \State $P_i \gets \texttt{FeasibleLocations}(i)$
        \State Determine class of pod $p$ (A/B/C)
        \If{Class A}
            \State Choose lowest-cost location
        \ElsIf{Class B}
            \State Choose second-best location
        \Else
            \State Choose third-best location
        \EndIf
        \State Assign pod $p$ to selected location
    \EndFor
    \State \Return $s''$
\EndFunction
\end{algorithmic}
\end{algorithm}

\subsubsection*{Lowest-Cost Repair}

This greedy heuristic always selects the feasible placement with the lowest cost for each destroyed iteration, independent of pod characteristics.

\begin{algorithm}[H]
\caption{Lowest-Cost Repair}
\begin{algorithmic}[1]
\Function{RepairLowestCost}{$s', I_{\text{destroy}}$}
    \For{$i \in I_{\text{destroy}}$}
        \State $P_i \gets \texttt{FeasibleLocations}(i)$
        \If{$P_i \neq \emptyset$}
            \State Assign pod in $i$ to lowest-cost location
        \EndIf
    \EndFor
    \State \Return $s''$
\EndFunction
\end{algorithmic}
\end{algorithm}

\subsubsection*{Random Repair}

To maintain solution diversity, this heuristic randomly selects a feasible placement for each destroyed iteration. It is especially helpful for escaping local optima.

\begin{algorithm}[H]
\caption{Random Repair}
\begin{algorithmic}[1]
\Function{RepairRandom}{$s', I_{\text{destroy}}$}
    \For{$i \in I_{\text{destroy}}$}
        \State $P_i \gets \texttt{FeasibleLocations}(i)$
        \If{$P_i \neq \emptyset$}
            \State Randomly assign pod to a location in $P_i$
        \EndIf
    \EndFor
    \State \Return $s''$
\EndFunction
\end{algorithmic}
\end{algorithm}
\subsection{Feasibility Checks}

Feasibility during repair is evaluated by three interdependent functions:
\begin{itemize}
    \item \textbf{RecallPlaces:} This function simulates the warehouse state at a given iteration and tracks pod locations. It's especially important for evaluating feasible placements in the future and finding previous positions of a target pod. 
    \item \textbf{IsPlaceFeasible:} This function checks if a place is feasible for storing a pod, i.e., it's unoccupied during the relevant time interval.
    \item \textbf{FeasibleLocations:} This function evaluates all storage locations to identify those that are feasible for a pod returning at a certain iteration. It uses RecallPlaces to simulate warehouse state and IsPlaceFeasible to validate placements. It also estimates cost (including transit cost if applicable). 
\end{itemize}
\begin{algorithm}[H]
\caption{RecallPlaces}
\begin{algorithmic}[1]
\Function{RecallPlaces}{$\text{solution}, \text{iteration}, \text{arrivals}, \text{departures}, \text{target\_pod}$}
    \State Initialize pod placements and station queues
    \State $\text{prev\_location} \gets \text{None}$
    \For{$x = 0$ to $\text{iteration}$}
        \State Retrieve pod arriving and departing at $x$
        \State Update warehouse configuration: remove departing pod, add arriving pod
        \If{$\text{target\_pod}$ is the one departing at $x$}
            \State $\text{prev\_location} \gets$ recorded location before departure
        \EndIf
        \State Update Station1 and Station2 based on queue dynamics
    \EndFor
    \State \Return current warehouse configuration at iteration, $\text{prev\_location}$, Station1, Station2
\EndFunction
\end{algorithmic}
\end{algorithm}
\begin{algorithm}[H]
\caption{IsPlaceFeasible}
\begin{algorithmic}[1]
\Function{IsPlaceFeasible}{$\text{config}, \text{solution}, \text{iteration}, \text{place\_id}, \text{next\_departure}, \text{destroy\_set}, \text{departures}$}
    \If{place is currently occupied in $\text{config}$}
        \State \Return False
    \EndIf
    \If{next departure exists}
        \For{each future iteration $j$ before $\text{next\_departure} - 1$}
            \If{another pod is assigned to $\text{place\_id}$ at $j$ and $j \notin \text{destroy\_set}$}
                \State \Return False
            \EndIf
        \EndFor
    \Else
        \For{future iterations $j$}
            \If{another pod is assigned to $\text{place\_id}$ and $j \notin \text{destroy\_set}$}
                \State \Return False
            \EndIf
        \EndFor
    \EndIf
    \State \Return True
\EndFunction
\end{algorithmic}
\end{algorithm}
\begin{algorithm}[H]
\caption{FeasibleLocations}
\begin{algorithmic}[1]
\Function{FeasibleLocations}{$\text{iteration}, \text{solution}, I_{\text{destroy}}, \text{arrivals}, \text{departures}, \text{warehouse}$}
    \State Get returning pod $p$ and origin station $s_{\text{from}}$ from departure at iteration
    \State Get next departure of $p$ as $t_{\text{next}}$
    \State $(\text{config}, \text{prev\_loc}, \_) \gets \texttt{RecallPlaces}(\text{solution}, \text{iteration}, \text{arrivals}, \text{departures}, p)$
    \State Initialize empty list $\text{feasible\_placements} \gets []$
    \For{each storage place $q$ in warehouse}
        \If{\texttt{IsPlaceFeasible}($\text{config}, \text{solution}, \text{iteration}, q, t_{\text{next}}, I_{\text{destroy}}, \text{departures}$)}
            \State Compute base cost: $c \gets c^{\text{from}}(s_{\text{from}}, q)$
            \If{$t_{\text{next}}$ exists}
                \State $c \gets c + c^{\text{to}}(q, s_{\text{next}})$
            \EndIf
            \State Append $(q, c)$ to $\text{feasible\_placements}$
        \EndIf
    \EndFor
    \State \Return $\text{feasible\_placements}$ and $\text{prev\_loc}$
\EndFunction
\end{algorithmic}
\end{algorithm}

\subsection{Acceptance Criteria}

We adopt Simulated Annealing (SA) as the acceptance criterion. A new solution $s''$ is always accepted if it improves the current best. Otherwise, it is accepted probabilistically:

\[
P(\text{accept}) = 
\begin{cases}
1 & \text{if } \text{cost}(s'') < \text{cost}(s) \\
\exp\left(-\frac{\Delta}{T}\right) & \text{otherwise}
\end{cases}
\]

where $\Delta = \text{cost}(s'') - \text{cost}(s)$ and $T$ is the current temperature. This strategy ensures a balance between exploration and exploitation throughout the search.

\section{Deep Reinforcement Learning for ALNS}

Traditional ALNS relies on weight-based selection of destroy and repair heuristics, where weights are updated based on historical performance (as shown in Section 3). While effective, this method does not explicitly incorporate the current search context or problem dynamics. 
To enhance the performance of ALNS in complex and dynamic settings, we incorporate a Deep Reinforcement Learning (DRL) agent (introduced by Reijnen et al.\cite{Reijnen2024}) to control the adaptive selection of operators and parameters during the search process. This hybrid approach, referred to as DR-ALNS, leverages learning-based decision-making to guide destroy and repair operations as well as acceptance parameters in an online manner. DR-ALNS models the ALNS configuration process as a Markov Decision Process (MDP), allowing a DRL agent to make context-aware decisions at each iteration.

Our implementation follows the general DR-ALNS framework proposed by Reijnen et al.~\cite{Reijnen2024}, adapted specifically to the pod repositioning context by integrating domain-specific components such as warehouse simulation and feasibility-aware placement evaluation.

\subsection{MDP Formulation}

Our implementation casts ALNS for pod repositioning as a Markov Decision Process with the following elements:

\begin{itemize}
  \item \textbf{State} \(s_t\): a fixed‐length vector comprising
    \begin{enumerate}
      \item Normalized temperature \(\;T/T_{\text{start}}\).
      \item Previous cost change \(\Delta_{t-1}/C^{\text{best}}\).
      \item Current gap to best \(\,(C^{\text{curr}} - C^{\text{best}})/C^{\text{best}}\).
      \item Normalized destroy‐operator weights (one entry per destroy heuristic).
      \item Normalized repair‐operator weights (one entry per repair heuristic).
      \item Current cost ratio \(C^{\text{curr}}/C^{\text{best}}\).
      \item Absolute best cost \(C^{\text{best}}\).
      \item Cost gap.
      \item Progress fraction \(t/T_{\max}\).
    \end{enumerate}
    Together, these components inform the agent about how “hot” the annealer is, how near the current solution is to the global best, which heuristics have shown promise, and how much of the search budget remains.
    
  \item \textbf{Action} \(a_t\): a single integer encoding a triple \((d_t,\;r_t,\;\delta_t)\), where
    \begin{itemize}
      \item \(d_t \in \{\text{destroy heuristics}\}\),
      \item \(r_t \in \{\text{repair heuristics}\}\),
      \item \(\delta_t \in \{\text{DoD levels}\}\) “DoD” stands for “Degree of Destruction”. In the context of ALNS, it specifies how much of the current solution is removed during the destroy phase.
    \end{itemize}
    Decoding into \((d_t,r_t,\delta_t)\) identifies which destroy operator to apply, which repair operator to follow, and how many pods to remove in that iteration.

  \item \textbf{Reward} \(r_t\): computed immediately after each destroy–repair iteration combined with simulated‐annealing acceptance. It consists of:
    \begin{enumerate}
      \item Normalized cost improvement \(\bigl(\Delta_t / C^{\text{init}}\bigr)\).
      \item New‐best bonus \(+1.0\) if the accepted solution establishes a new global best.
      \item Fluctuation penalty \(-0.5\) if the last three cost deltas follow “down–up–down.”
      \item Repair‐failure penalty \(-0.2\) if the repair step was infeasible.
      \item Rejection penalty \(-0.1\) if simulated annealing rejects the candidate.
      \item Exploration bonus \(+0.1 \times (T/T_{\text{start}})\) if a non‐improving but feasible solution is accepted.
    \end{enumerate}
    This composite reward encourages steady cost reduction, discourages zig-zag behavior, penalizes infeasibility and rejections, and lightly rewards early exploration.

  \item \textbf{Transition} \(\bigl(s_t, a_t\bigr)\to\bigl(s_{t+1}, r_t\bigr)\):
    \begin{enumerate}
      \item Decode: the integer \(a_t\) into a chosen destroy operator \(d_t\), repair operator \(r_t\), and DoD level \(\delta_t\).
      \item Destroy phase: remove \(\delta_t\) pods from the current solution using \(d_t\). If no pods can be removed, immediately assign \(r_t = -1.0\) and keep \(s_{t+1} = s_t\).
      \item Repair phase: rebuild a full solution with \(r_t\), yielding cost difference \(\Delta_t = C^{\text{old}} - C^{\text{new}}\) and a failure indicator.
      \item Simulated‐annealing acceptance: accept unconditionally if \(\Delta_t > 0\); otherwise accept with probability \(\exp\bigl(\Delta_t/T\bigr)\). If accepted and \(C^{\text{new}} < C^{\text{best}}\), update the global best.
      \item Operator‐weight update: assign a score \(\sigma \in \{0,1,2,3\}\) to both \(d_t\) and \(r_t\) (0 for failure/rejection, 1 if accepted without improvement, 2 if accepted with improvement but not new-best, 3 if new-best). Then update each selected weight via
      \[
        w \;\leftarrow\; 0.95\,w \;+\; 0.05\,\sigma,\quad w \ge 0.1.
      \]
      Record \(\Delta_t\) in a length-3 buffer for possible fluctuation penalty.
      \item Reward computation: compute \(r_t\) as defined above.
      \item Temperature cooling: \(T \leftarrow \max(T_{\text{stop}},\,T \times \text{decrease\_factor})\).
      \item Increment iteration: \(t \leftarrow t + 1\).
      \item Termination check: stop if \(T \le T_{\text{stop}}\) or \(t \ge T_{\max}\).
      \item Next observation: \(s_{t+1}\) is constructed using the same nine‐element structure.
    \end{enumerate}

  \item \textbf{Policy} \(\pi_\theta\): a parameterized mapping (e.g., via Proximal Policy Optimization) from each observed \(s_t\) to a distribution over the discrete actions \(\{0,\dots,|\mathcal{D}|\times|\mathcal{R}|\times|\Delta|-1\}\). Over training episodes, \(\pi_\theta\) learns which destroy/repair/DoD combinations yield the largest cumulative reward (i.e., the lowest repositioning cost).
\end{itemize}

\subsection{DRL Integration in ALNS for the Pod Repositioning Problem}

Building on the MDP definition, DRL integration proceeds as follows:

\begin{enumerate}
  \item \textbf{Composite Action Selection.}  
    At each iteration, the agent samples \(a_t\) from \(\pi_\theta(\cdot \mid s_t)\), simultaneously choosing a destroy operator, a repair operator, and a DoD level. This replaces the traditional weight‐based roulette selection with a learned policy that directly accounts for the current search context.

  \item \textbf{Online Parameter Adaptation.}  
    By observing normalized operator weights in each \(s_t\), the policy can bias toward heuristics that have performed well recently, while retaining the capacity to explore under‐utilized options thanks to the enforced lower bound on weights. The annealing temperature \(T\) and the iteration index \(t\) further inform the policy about exploration–exploitation trade‐offs.

  \item \textbf{Reward‐Driven Learning.}  
    The shaped reward balances immediate cost improvements, sparse global‐best discoveries, penalties for unproductive patterns, and exploration bonuses. This encourages the agent to discover long‐term strategies (e.g., applying a heavy destruction early, switching heuristics mid‐search, or gradually tightening acceptance criteria) rather than solely focusing on short‐term gains.

  \item \textbf{Simulated‐Annealing Mechanism.}  
    The MDP’s transition incorporates an SA acceptance rule: if \(\Delta_t > 0\), accept; otherwise accept with probability \(\exp\bigl(\Delta_t/T\bigr)\). This stochastic element preserves exploration capabilities even when the policy proposes non‐improving moves, especially at higher temperatures. Over time, as \(T\) decays, the agent’s decisions become more greedy.

  \item \textbf{Training Loop.}  
    An off‐the‐shelf DRL algorithm (e.g., Proximal Policy Optimization) interacts with the MDP:
    \begin{itemize}
      \item Observe \(s_t\).
      \item Sample \(a_t \sim \pi_\theta(\cdot \mid s_t)\).
      \item Execute the combined destroy–repair–SA step, receive \(r_t\) and \(s_{t+1}\).
      \item Store the transition \((s_t, a_t, r_t, s_{t+1}, \text{done})\).
      \item Periodically update \(\theta\) to maximize expected discounted return.
    \end{itemize}
    Over many episodes—each consisting of up to \(T_{\max}\) destroy–repair iterations—the policy learns which operator sequences and DoD choices yield the best repositioning cost reductions in varied search contexts.
\end{enumerate}

By combining composite action selection, operator‐weight feedback, SA acceptance, and tailored reward shaping, this framework realizes a lightweight, efficient DR-ALNS approach for the Pod Repositioning Problem.

\section{Computational Results}
This section first describes the test instances, then outlines the parameter learning process for both ALNS and DR\_ALNS, and concludes with a computational comparison against all baseline methods.

\subsection{Instance Description}

To evaluate the performance of our proposed ALNS algorithm, we adopt two benchmark instances introduced by Krenzler et al.~\cite{Krenzler2018}, which model pod repositioning within an RMFS. These instances are well-suited for assessing both the effectiveness and scalability of pod placement strategies under realistic warehouse conditions.

The \textit{small instance} comprises a storage area with 10 locations and 10 pods, operating over 1{,}000 discrete time steps. The warehouse layout is one-dimensional and symmetric with respect to two identical picking stations. This configuration allows for clear visualization of pod dynamics and cost patterns, making it ideal for qualitative algorithm analysis and debugging.

The \textit{medium instance} represents a significantly larger and more complex environment, featuring 504 storage locations, 441 pods, and two asymmetric picking stations. The system is simulated over 20{,}000 time steps, during which approximately 20{,}000 repositioning decisions are made---excluding a short initialization period required to fill the station queues. Pod departures follow a weighted stochastic process that favors high-frequency pods and prioritizes one of the stations with higher throughput, thereby emulating non-uniform demand patterns observed in real-world e-commerce warehouses.

Both instances are based on a deterministic model in which pod movements incur cost according to Manhattan distances between storage locations and pick stations. The problem setting strictly adheres to passive repositioning: a pod is only moved after serving a pick station, with no predictive or anticipatory relocation. This setup enables direct comparison with the baseline heuristics proposed in Krenzler et al.~\cite{Krenzler2018}, including \textit{Cheapest Place}, \textit{Fixed Place}, \textit{Fixed Place (Approximate)}, \textit{Random Place}, \textit{Binary Integer Programming}, \textit{BIP-Iterative}, \textit{Genetic Algorithms}, and the \textit{Tetris heuristic}. \textit{Cheapest Place} selects the nearest available location; \textit{Fixed Place} assigns each pod a dedicated storage slot; \textit{Fixed Place (Approximate)} uses usage frequency to assign pods to ranked zones; \textit{Random Place} chooses locations randomly; \textit{Binary Integer Programming} (BIP) offers an exact, optimization-based solution over the full time horizon; \textit{BIP-Iterative} solves smaller subproblems sequentially to handle scalability; \textit{Genetic Algorithms} use evolutionary search to improve placements; and the \textit{Tetris heuristic} prioritizes frequently used pods for optimal positions based on occupation intervals.

\subsection{Parameter Learning for ALNS and DR\_ALNS}
All experiments of ALNS and DR\_ALNS were conducted on a workstation equipped with an Intel Core i9-10850K CPU @ 3.60\,GHz, 32~GB RAM, and an NVIDIA RTX~3060 GPU.

\subsection*{ALNS: Simulation-Based Parameter Optimization}
To ensure a fair comparison with the DR\_ALNS, we employ a simulation-based procedure to calibrate the parameters of the ALNS and generate a high-quality reference solution derived exclusively from the ALNS framework. The tuning process focuses on four main parameters of the algorithm:

\begin{itemize}
    \item \textbf{$T_{\text{start}}$}: Initial temperature for the simulated annealing acceptance criterion.
    \item \textbf{$T_{\text{stop}}$}: Final temperature that determines when cooling ends.
    \item \textbf{MLength}: Markov chain length, i.e., the number of iterations at a fixed temperature.
    \item \textbf{DecreaseFactor}: Cooling rate $\alpha$ that controls how quickly $T_{\text{start}}$ is reduced.
\end{itemize}

The parameter ranges were selected based on preliminary tests and expert judgment:
\begin{align*}
T_{\text{start}} &\in \{5, 7.5, 10, 12.5, 15\}, \\
T_{\text{stop}} &\in \{0.1, 0.3, 0.5, 0.7, 0.9\}, \\
\text{MLength} &\in \{30, 40, 50\}, \\
\text{DecreaseFactor} &\in \{0.76, 0.80, 0.84, 0.88, 0.92, 0.95\}.
\end{align*}

Fifty random parameter combinations were tested in the small instance. Each was evaluated by running ALNS and recording the total cost and runtime. To encourage exploration, the acceptance criterion included a $25\%$ chance of accepting worse solutions. Additionally, when no improvement occurred for several consecutive iterations, a reinitialization of parameters was triggered to escape potential local optima.

The best configuration found was:
\[
T_{\text{start}} = 12.5, \quad T_{\text{stop}} = 0.1, \quad \text{MLength} = 30, \quad \text{DecreaseFactor} = 0.95
\]

This configuration was used in the final experimental evaluation due to its superior balance between solution quality and runtime.
\subsection*{DR\_ALNS: Policy Training}

Policy learning was conducted exclusively on the small instance. A total of 20{,}000 training timesteps were used, with each episode comprising 1{,}000 ALNS iterations. Training was performed in Python~3.11.4 with Stable Baselines3 (v2.5.0), and required approximately 1.1 hours to complete. The resulting PPO agent was saved as a model checkpoint for subsequent evaluation and inference. The main PPO hyperparameters were as follows: a learning rate of $1 \times 10^{-3}$, batch size of 128, $n_\text{steps} = 2{,}048$, and an entropy coefficient of 0.01 to encourage exploration.

For inference, we evaluate the trained DR-ALNS PPO agent in the same warehouse environment configuration used during training. The inference environment employs 1,000 ALNS iterations, an initial temperature $T_{\text{start}}$ of 1.0, a stopping temperature $T_{\text{stop}}$ of 0.001, and a temperature decrease factor of 0.98. 
\subsection{Comparison with baselines}

\begin{figure}[h!]
    \centering    \includegraphics[width=0.59\linewidth]{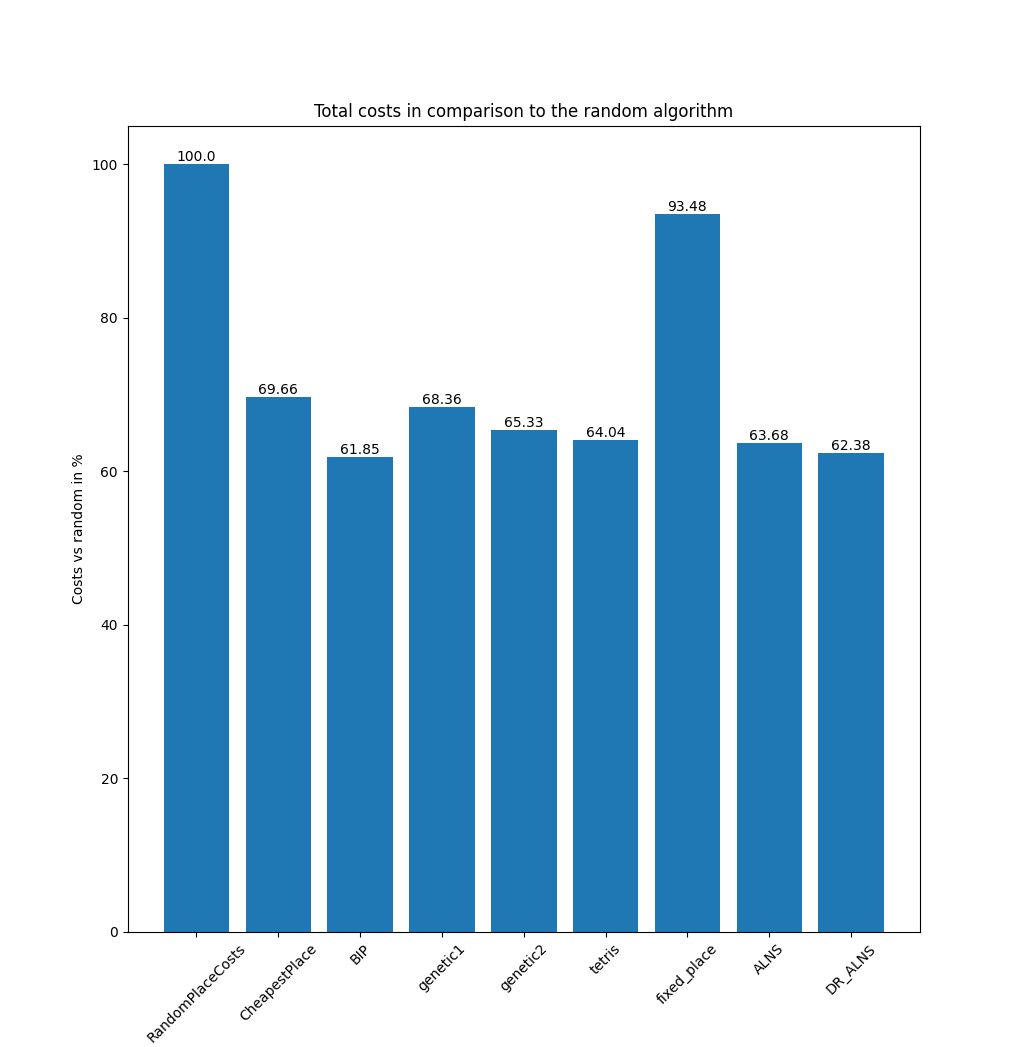}
    \caption{Comparison of ALNS, DR\_ALNS with baseline methods for the small instance.}
    \label{fig:result_small}
\end{figure}

\begin{figure}[h!]
    \centering
    \includegraphics[width=0.59\linewidth]{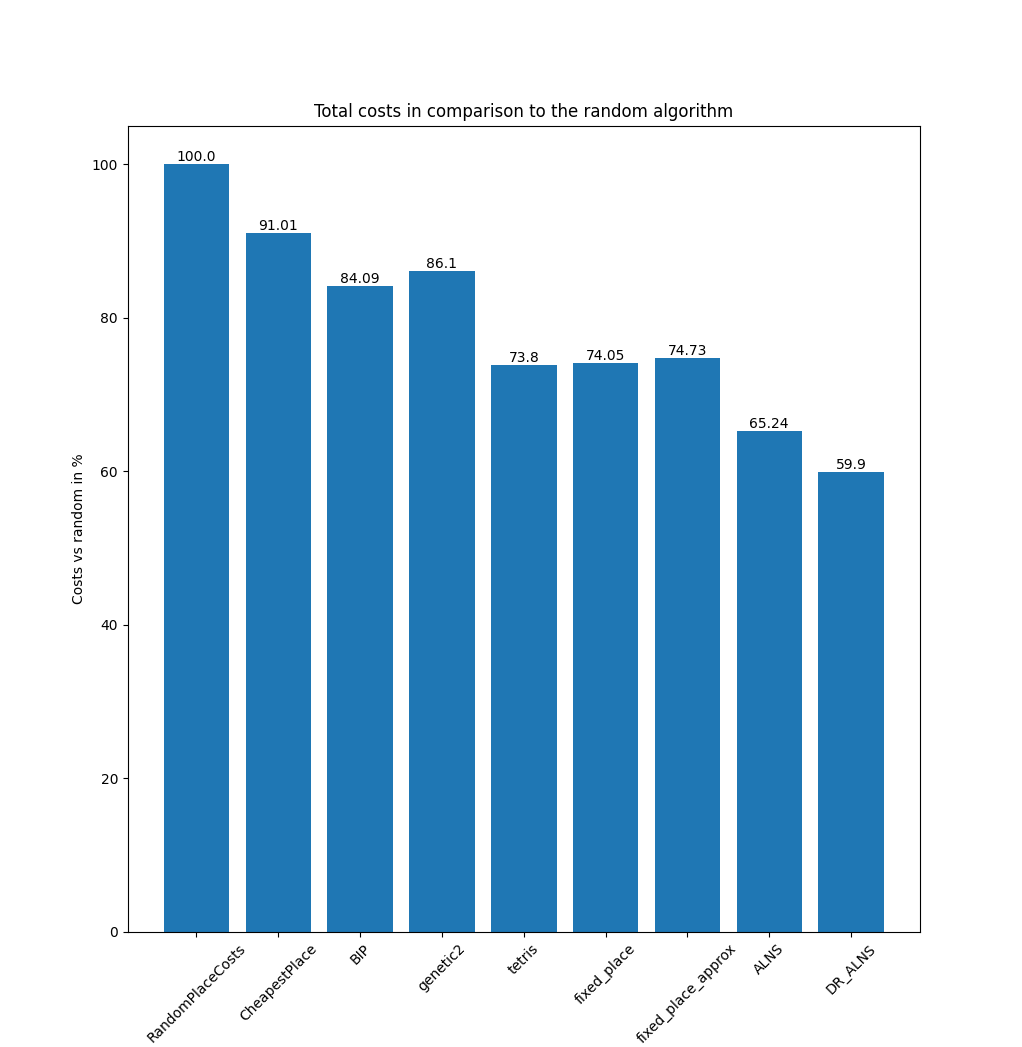}
    \caption{Comparison of ALNS, DR\_ALNS with baseline methods for the medium instance.}
    \label{fig:result_medium}
\end{figure}

Figures~\ref{fig:result_small} and \ref{fig:result_medium} present the total costs of various heuristics as a percentage relative to the random placement baseline (100\%), evaluated on small and medium warehouse instances, respectively.

On the \textbf{small instance} (Figure~\ref{fig:result_small}), \textit{BIP}, which computes the optimal solution, is the top performer on the \textbf{small instance} at \textbf{61.85\%}, closely followed by \textit{DR\_ALNS} (62.38\%) and \textit{ALNS} (63.68\%). Classic heuristics such as \textit{Cheapest Place} and \textit{Tetris} remain competitive, while \textit{Fixed Place} incurs significantly higher costs (93.48\%), indicating limited adaptability in smaller-scale settings.

On the \textbf{medium instance} (Figure~\ref{fig:result_medium}), \textit{DR\_ALNS} achieves the best result with a cost of \textbf{59.9\%}, followed by \textit{ALNS} (65.24\%) and \textit{Tetris} (73.8\%). Traditional heuristics such as \textit{Cheapest Place} (91.01\%) and \textit{Fixed Place} variants (74.75\%) show noticeably higher costs.

Overall, \textit{DR\_ALNS} consistently ranks among the best-performing methods across both instance sizes, highlighting the benefits of learning-based and online adaptive control. In contrast, traditional heuristics exhibit reduced effectiveness, particularly as problem complexity increases.

In terms of efficiency, all baseline methods—except for \textit{BIP} (on the medium instance) and the \textit{Genetic Algorithm}—complete within one minute for both instance sizes. \textit{ALNS} solves the small instance in 103 seconds and the medium instance in about 27 minutes. The inference times of \textit{DR\_ALNS} are 74 and 400 seconds for the small and medium instances, respectively.

Notably, both \textit{ALNS} and \textit{DR\_ALNS} maintain strong performance on the medium instance using parameters tuned only on the small instance (via simulation optimization for ALNS and policy training for DR\_ALNS). This demonstrates good generalization capability and robustness—especially for \textit{DR\_ALNS}, which adapts effectively to larger problem scales without the need for re-tuning.

\vspace{-0.5cm}
\section{Conclusion and Outlook}

In this work, we adapted the DR-ALNS framework to address the deterministic Pod Repositioning Problem (PRP), demonstrating its effectiveness in structured warehouse environments. By incorporating domain-specific destroy and repair heuristics, we enhanced the flexibility of the ALNS approach to better match the operational characteristics of the PRP. Computational experiments across different instance sizes show that DR-ALNS consistently delivers comparable or superior performance to classical baselines such as \textit{Cheapest Place}, \textit{Fixed Place}, \textit{Tetris}, and \textit{Binary Integer Programming}, achieving both high solution quality and robustness. Notably, the policy used in DR-ALNS was trained solely on the small instance and successfully generalized to larger instances without re-training, highlighting the adaptability and transferability of the learned policy across problem scales.

As a next step, future work could explore extending the approach to stochastic or real-time settings, or integrating it with other warehouse decision-making layers such as order picking or inventory management, to further enhance its practical applicability.

\begin{credits}
\subsubsection{\ackname} I would like to thank Marco Ochoa and Lisa van Oost, students in my Warehousing course at the University of Twente, for their contributions to the initial design and implementation ideas of the ALNS approach.
\end{credits}
%
%
%
\bibliographystyle{splncs04} \bibliography{bib}
\end{document}